\documentclass[journal]{IEEEtran}
\pdfoutput=1
\usepackage{authblk}
\usepackage{fullpage}
\usepackage{amssymb,amsmath}
\usepackage[utf8x]{inputenc}
\usepackage[T1]{fontenc}
\usepackage{siunitx}
\usepackage[version=3]{mhchem}
\usepackage{float}
\usepackage{cite}
\usepackage{url}

\usepackage[left]{lineno}

\usepackage{setspace}

\usepackage{graphicx}
\graphicspath{{figures/}}
\begin{document}
\title{Unsupervised High Impedance Fault Detection Using Autoencoder and Principal Component Analysis}

\author{{Yingxiang~Liu,
        Mohammad~Razeghi-Jahromi
        and~James~Stoupis}
\thanks{Y. Liu is with the Ming Hsieh
Department of Electrical and Computer Engineering
, University of Southern California, Los Angeles, CA, 90089 USA (e-mail: yingxian@usc.edu)}
\thanks{M.Razeghi-Jahromi is with ABB Corporate Research United
States (USCRC), Raleigh, NC 27606 USA (e-mail: mohammad.razeghijahromi@us.abb.com)}
\thanks{J.Stoupis is with ABB Corporate Research United
States (USCRC), Raleigh, NC 27606 USA (e-mail: james.stoupis@us.abb.com)}}


\maketitle

\begin{abstract}
Detection of high impedance faults (HIF) has been one of the biggest challenges in the power distribution network. The low current magnitude and diverse characteristics of HIFs make them difficult to be detected by over-current relays. Recently, data-driven methods based on machine learning models are gaining popularity in HIF detection due to their capability to learn complex patterns from data. Most machine learning-based detection methods adopt supervised learning techniques to distinguish HIFs from normal load conditions by performing classifications, which rely on a large amount of data collected during HIF. However, measurements of HIF are difficult to acquire in the real world. As a result, the reliability and generalization of the classification methods are limited when the load profiles and faults are not present in the training data. Consequently, this paper proposes an unsupervised HIF detection framework using the autoencoder and principal component analysis-based monitoring techniques. The proposed fault detection method detects the HIF by monitoring the changes in correlation structure within the current waveforms that are different from the normal loads. The performance of the proposed HIF detection method is tested using real data collected from a 4.16 kV distribution system and compared with results from a commercially available solution for HIF detection. The numerical results demonstrate that the proposed method outperforms the commercially available HIF detection technique while maintaining high security by not falsely detecting during load conditions.
\end{abstract}


\IEEEpeerreviewmaketitle

\section{Introduction}\label{sec:Intro}
\IEEEPARstart{H}{igh} impedance fault (HIF) is a group of power system disturbances that typically occurs when a live conductor contacts a surface with high impedance. The HIF current magnitude typically ranges from 0 to 75 A, and the characteristics of HIFs are affected by various factors such as surface type and load conditions \cite{ ghaderi:2017}. The low current magnitudes and diverse characteristics make the HIFs difficult to be detected using conventional over-current relays \cite{wester:1998}. It is estimated that between 5\% and 10\% of the distribution faults are HIF \cite{ adamiak:2006}, and about 25\% of the HIFs are not detected using the over-current relays \cite{ russell:1995}. Since over-current relays usually cannot detect HIFs, the arcs and flashover caused by HIFs can cause fires and jeopardize human safety \cite{chaitanya:2020}. Therefore, effectively detecting HIFs remains a non-negligible challenge.

Recent advances in the industrial internet of things and smart grid allow increasing computation resources and data analysis capabilities within the power grid \cite{feng_wang:2021, qiu_chi:2020}. As a result, machine learning approaches have been gaining popularity for HIF detection. Ghaderi et al. \cite{ ghaderi:2015} trained a support vector machine (SVM) classifier with features of current waveform energy and normalized joint time-frequency moments. Baqui et al. \cite{ baqui:2011} combined artificial neural network (ANN) with discrete wavelet transforms (DWT) for HIF detection in medium-voltage networks. Features were extracted from current measurements using DWTs and then fed into the ANN for classification. Wang et al. \cite{ wang_dehghanian:2020} first applied a modified Gabor WT to the input signal to extract two-dimensional scalograms and then applied a two-dimensional convolutional neural network (CNN) for classification. In \cite{ veerasamy:2021}, a Long Short Term Memory (LSTM) classifier was trained with features obtained from DWT analysis to detect the HIFs in the solar photovoltaic integrated power system.

The machine learning-based studies above use supervised learning methods to detect HIF by training classification models that map the input to a set of labels corresponding to different HIF types. However, there are some limitations to using classification for HIF detection. The first one is the generalization problem of the models. The supervised learning-based HIF detection methods detect the occurrence of HIF by performing classification using models trained with labeled data collected under various normal load conditions and during different HIFs. However, when the supervised HIF detection methods are deployed in the grid, the classifiers may produce undependable results if the load profile or HIFs are not present in the training set. Another limitation is scalability. Since the HIF detection method needs to be deployed to different parts of the grids with various load profiles, the model needs to be trained with data collected from different utilities from different parts of the grid to ensure the data-driven model works for all of them. Therefore, the supervised learning methods are different to scale in real-world applications. To deal with the limitations mentioned above, the unsupervised learning methods can be used for fault detection since they do not require labels and can easily adapt to different load conditions. The fault detection methods based on unsupervised learning methods have been successfully applied to various engineering applications such as chemical and semiconductor manufacturing \cite{ qin:2003}. However, their applications for HIF detection are still limited. In recent years, Rai et al. \cite{rai:2021} applied a convolutional autoencoder trained with simulated HIF scenarios. Then cross-correlation between the reconstructed signal and the original signal was used to discriminate HIFs from loads. Although the proposed method showed good fault detection performance on the simulated dataset, it relies on training using faulty HIF data, which is difficult to acquire in real-world applications. Sarwar et al. \cite{sarwar:2020} introduced principal component analysis-based statistical process monitoring techniques to detect HIF. The proposed methods can successfully detect the occurrence of HIF. However, instead of analyzing the measurements collected from one location in the grid, it applies PCA to 29 variables simulated from the IEEE 13-node test feeder. As a result, it requires resource-intensive communication and data storage between multiple measurement devices.

Consequently, this paper proposed an unsupervised HIF detection framework based on the autoencoder (AE) and principal component analysis, which are trained using historical measurements collected from one location in the grid. First, the univariate current measurement is augmented into a data matrix consisting of multiple variables. Then the autoencoder extracts nonlinear features from the data matrix to capture the correlations among different variables. Next, a PCA model is built based on the autoencoder's reconstruction errors. Finally, the PCA-based statistical monitoring technique is used to characterize the residuals from the AE model of the normal load data and establish thresholds based on various statistics. The autoencoder and the PCA can then be deployed online to monitor the new current measurement. If HIF occurs, the correlation structure of the augmented data matrix will deviate from the correlation learned by the AE from the normal loads, thus leading to abnormal reconstruction errors from the autoencoder and reflected in indices of the PCA-based monitoring model. The main contributions of this study are: (1) combine autoencoder and PCA model to characterize the correlations structure of univariate current measurement; (2) introduce statistical process monitoring techniques for detecting HIF using data collected from a single location in the grid; and (3) The proposed unsupervised method only relies on the measurements of the normal loads. In addition, since the number of parameters in the AE model is small, the proposed method can be trained rapidly and thus can be easily adapted and deployed to computing devices located across the grid. The remainder of this paper is organized as follows. Section \ref{sec:backgroud} introduces autoencoder and PCA-based process monitoring technique, followed by the details of the proposed HIF detection method in Section \ref{sec:method}. Section \ref{sec:result} presents a real dataset collected from a 4.16 kV distribution system to evaluate the effectiveness of the proposed method. Finally, the conclusions are presented in Section \ref{sec:conclusion}.

\section{Preliminaries} \label{sec:backgroud}
\subsection{Autoencoder}
Autoencoder is an unsupervised neural network that learns to compress and reconstruct the input data effectively. It has been widely used for fault detection in various applications such as electric motors \cite{ principi:2019}, wind turbines \cite{ jiang:2017}, and chemical processes \cite{ chen:2020}. An autoencoder consists of two parts: an encoder followed by a decoder which can be represented using different neural network structures such as multi-layer perceptron (MLP), convolutional neural network (CNN), and recurrent neural network (RNN). In this study, we used the MLP as the encoder and decoder due to its simplicity. For an autoencoder composed of a single hidden layer, the encoder maps the input vector $\mathbf{x}\in\mathbb{R}^M$ in the hidden representation $\mathbf{h}\in\mathbb{R}^P$ as follows.
\begin{equation}
    \mathbf{h} = f(\mathbf{W}_1\mathbf{x}+ \mathbf{b}_1)
\end{equation}
where $f$ is an non-linear activation function, $\mathbf{W}_1\in\mathbb{R}^{P\times M}$ is a weight matrix, and $\mathbf{b}_1 \in\mathbb{R}^{P}$ is a bias vector.
The decoder then tries to reconstruct the input $\mathbf{x}$ by
using 
\begin{equation}
    \Tilde{\mathbf{x}} = f(\mathbf{W}_2\mathbf{h}+ \mathbf{b}_2)
\end{equation}
where $\mathbf{W}_2\in\mathbb{R}^{M\times P}$ is the decoder weight matrix, and $\mathbf{b}_2 \in\mathbb{R}^{M}$ is the bias vector, and $\Tilde{\mathbf{x}}$ is the reconstructed input vector. To avoid the autoencoder learning to copy the input to the output and to capture the correlation among different input variables, the dimension of the hidden layer $\mathbf{h}$ is chosen to be smaller than the dimension of the input. Training of the autoencoder is performed by minimizing the mean squared error (MSE) loss function:
\begin{equation} \label{eq:loss}
    L(\mathbf{\theta}) = ||\mathbf{x} - \Tilde{\mathbf{x}}||^2
\end{equation}
where $\mathbf{\theta}$ represents all the network parameters.

\subsection{PCA for Fault Detection}\label{ssec:PCA_fault}
Principal Component Analysis (PCA) is widely used as a dimensional reduction tool in different domains such as computer science and electrical engineering \cite{duan:2021, ma_yuan:2019, rafferty:2016}. It produces a low-dimensional representation of multivariate data by finding a direction or subspace of the largest variance in the original measurement space. Let $\mathbf{X}\in\mathbb{R}^{N\times M}$ denotes a data matrix with each row representing a sample $\mathbf{x}\in\mathbb{R}^M$. After applying PCA to the data matrix ${\mathbf{X}}$, it can be decomposed as,
\begin{equation}
    \mathbf{X} = \mathbf{T}\mathbf{P}^\top + \Tilde{\mathbf{T}}\Tilde{\mathbf{P}}^\top
\end{equation}
where $\mathbf{P}$ consists of the first $l$ loading vectors that contain most variance of the data and $\Tilde{\mathbf{P}}$ is the last $M-l$ loading vectors. The subspace spanned by $\mathbf{P}$ is known as the principal component subspace (PCS) and that spanned by $\Tilde{\mathbf{P}}$ is called the residual subspace (RS). Consequently, the measurement space can be divided into the PCS and the RS, where the PCS contains normal or major variations, and the RS contains small variations or noises.

PCA has been widely used for statistical process monitoring \cite{qin:2003, macgregor:marlin:kresta:skagerberg:1991, macgregor:1989} and fault detection of multivariate data collected from chemical processes. It is used to model the normal static variation from data related to  normal operation. To perform fault detection, the general idea is first to build models using data collected during normal operations. Then control limits are established to define normal operation regions. Finally, the models and the control limits are applied to new data for online fault detection.  With a PCA model, different fault detection indices such as Hotelling’s $T^2$ index, the SPE (or Q index) index and the combined index $\varphi$ can be defined to monitor various aspects of the data. It is important to note that these indices and the corresponding limits assume that the data samples are independent in time.
\begin{enumerate}
    \item Hotelling’s $T^2$ index \\
    Hotelling’s $T^2$ index measures variations in the PCS,
\begin{equation}\label{eq:T2_ind}
    T^2 = \mathbf{x}^\top\mathbf{P}\mathbf{\Lambda}^{-1}\mathbf{P}^\top\mathbf{x}
\end{equation}
where $\mathbf{\Lambda}$ is the convariance matrix of the latent scores matrix $\mathbf{T}$. It can be proven that $T^2$ statistic follows a $F$ distribution,
\begin{equation}
    \frac{N(N-l)}{l(N^2-1)}T^2 \sim F_{l,N-l} 
\end{equation}
where $F_{l,N-l}$ is an $F$ distribution with $l$ and $N-l$ degrees of freedom \cite{tracy:young:mason:1992}. As a result, for a given confidence level $\alpha$, the control limit can be calculated based on the $F_{l, N-l}$ distribution. The index is considered normal if 
\begin{equation}
    T^2 \leq T_{\alpha}^2 \equiv \frac{l(N^2 - 1)}{N(N-l)}F_{l,N-l;\alpha}
\end{equation}
If the number of data points $N$ is large, the $T^2$ index can be well approximated with a $\chi^2$ distribution with $l$ degrees of freedom \cite{qin:2003} and 
\begin{equation} \label{eq:T2_lim}
    T_{\alpha}^2 = \chi_{l;\alpha}^2
\end{equation}
The $T^2$ index measures the distance to the origin in the principal component subspace, which contains normal process variations with large variance. The variation of the projection of a sample vector $\mathbf{x}$ on the PCS is considered normal if its $T^2$ index is less than the control limit $T_{\alpha}^2$.
\item SPE (Squared Prediction Error) index \\
The SPE index measures the projection of a sample vector $\mathbf{x} \in \mathbb{R}^M$ onto the residual space. It is defined as the squared norm of the residual vector $\Tilde{\mathbf{x}}$.
\begin{equation}\label{eq:SPE_ind}
    \text{SPE}(\mathbf{x}) = ||\Tilde{\mathbf{x}}||^2 = \mathbf{x}^\top\Tilde{\mathbf{P}}\Tilde{\mathbf{P}}^\top\mathbf{x}
\end{equation}
The control limit of the SPE index can be derived using the result in \cite{box:1954},
\begin{equation} \label{eq:SPE_lim}
    \delta_\alpha^2 = g\chi_{h;\alpha}^2
\end{equation}
where
\begin{equation}
    g = \frac{\sum_{i=l+1}^{M}{\lambda_i^2}}{\sum_{i=l+1}^{M}{\lambda_i}}, \ 
    h = \frac{(\sum_{i=l+1}^{M}{\lambda_i})^2}{\sum_{i=l+1}^{M}{\lambda_i^2}}
\end{equation}
$\alpha$ is confidence level. $l$ is the number of PC in the principal component subspace, and $\lambda_i$ is the $i^{th}$ eigenvalue of the sample convariance matrix $\frac{1}{N-1}\mathbf{X}^\top\mathbf{X}$. 

Since the SPE index focuses on the residual subspace, it measures the variability that breaks the static process relations. If the SPE index is above the control limit $\delta_{\alpha}$, it indicates a fault occurs that breaks the normal correlation structure.
\item Combined index \\
If both the $T^2$ index and SPE index are equally important, a global index can be used to combine the two indices, such as the combined index $\varphi$ \cite{yue:qin:2001, dong:qin:2020DPM}. This results in monitoring one index instead of two. The combined index is defined as follows,
\begin{equation} \label{eq:combine_lim}
    \varphi = T^2(\mathbf{x}) + g^{-1}\text{SPE}(\mathbf{x}) \sim \chi_{l+h}^2
\end{equation}
where $g$ and $h$ come from the calculation of the SPE control limit. With $\alpha$ as the confidence level, the control limit of the combined index is $\chi_{l + h;\alpha}^2$. As a result, a fault is detected if the value of $\varphi$ is greater than the control limit.
\end{enumerate}
\section{Proposed HIF Detection Procedure} \label{sec:method}
\begin{figure}[h]
  \centering
  \includegraphics[width=0.9\linewidth]{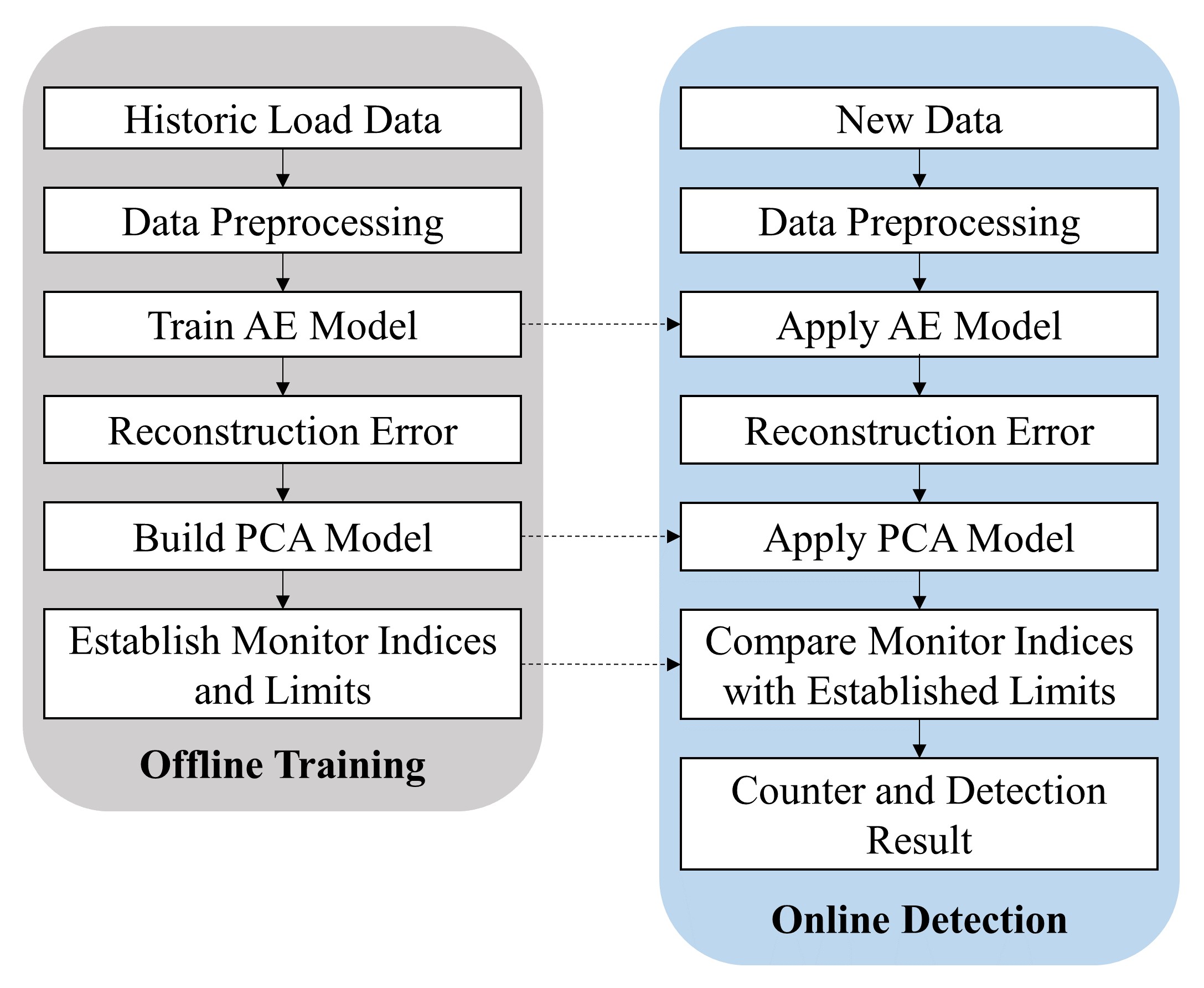}
\caption{Workflow of the proposed HIF detection method.}
\label{fig:flow}
\end{figure}

The occurrence of HIF introduces minor random distortions in current waveforms. As a result, the correlation between the current measurements between different cycles will show inconsistency from the correlation structure of the measurements collected during normal load conditions. Therefore, the proposed fault detection procedure detects the HIF by monitoring the changes in correlation structure within the current waveforms. The workflow of the proposed HIF detection is shown in Figure \ref{fig:flow}.
\subsection{Data Preprocessing}
The proposed HIF detection method first converts the single-phase current waveform to a data matrix by sampling at the same locations within each cycle across the historical measurement of loads. Let $ts$ be the number of samples per cycle and the length of the original signal to be $N \times ts$, the original signal can be represented as $S = [s_{(1)}, s_{(2)}, ... , s_{(N\times ts - 1)}, s_{(N\times ts)}]$. With $M$ to be the number of variables and $\Delta = ts / M$ be the gap when sampling from the original signal $S$, the matrix $\mathbf{X}$ can be written as \\
\begin{gather} 
\mathbf{X} = 
\begin{bmatrix}
s_{(1)} & s_{(1 + \Delta)} & ... & s_{(ts)}\\
s_{(1 + ts)} & s_{(1 + \Delta + ts)} & ... & s_{(2ts)}\\
... & ... & ... & ...
\end{bmatrix}
\end{gather}
The resulting matrix $\mathbf{X}$ has $M$ columns and $N$ rows. Since the autoencoder will be trained to reconstruct each row of the data matrix, the sampling is used to reduce the network’s input dimension and thus decrease the total number of parameters in the neural network model to prevent overfitting and improve training speed.
\subsection{Offline Training}
In the offline training step, an autoencoder and a PCA model are built to characterize the correlation structures of the current waveforms of normal loads. An autoencoder model is trained to extract the normal correlation and nonlinear features from the augmented data matrix by minimizing the MSE loss in Equation \ref{eq:loss}. After the autoencoder is trained to reconstruct the data matrix formed using normal load current waveforms, it can remove common features from the data matrix, leaving small residuals for all the variables in the data matrix. As a result, the autocorrelations within the input data matrix are eliminated, and the residuals only contain static variations, which can be modeled using the PCA and lend themselves to detect faults. PCA-based process monitoring techniques are applied to the reconstruction errors or the residuals of the fault-free data matrix produced by the trained autoencoder. Let $\Tilde{\mathbf{X}}$ be the output of the trained autoencoder. The reconstruction error of the data matrix can be written as,
\begin{equation} 
    \mathbf{E} = \mathbf{X} - \Tilde{\mathbf{X}}
\end{equation} 
After normalizing each column of $\mathbf{E}$ to have zero mean and unit variance, a PCA model can be built from the normalized reconstruction error. Then the number of latent variables $l$ can be selected based on cumulative percent variance (CPV)
\begin{equation}
    CPV(l) = \frac{\sum_{i=1}^{l}\lambda_i}{\sum_{i=1}^{M}\lambda_i}
\end{equation}
With the selected $l$ and confidence level $\alpha$, the control limits for SPE, $T^2$, and $\varphi$ indices can be established using Equations \ref{eq:SPE_lim}, \ref{eq:T2_lim}, and \ref{eq:combine_lim}.
\subsection{Online HIF Detection}
The trained autoencoder and PCA model are applied to three phases separately for detecting the high impedance fault in new measurements. For each phase, after acquiring the new current measurement of a cycle, a vector $\mathbf{x}$ with $M$ variables is constructed by sampling from the cycle. Then the new vector is passed as an input to the trained autoencoder model to get a vector of reconstruction errors $\mathbf{e}\in\mathbb{R}^{M}$. Since the autoencoder is trained using data from normal loads, abnormal reconstruction errors of the vector can be observed if the occurrence of HIF distorts the correlation structure within a cycle. After scaling reconstruction errors $\mathbf{e}$ with the mean and variances calculated when building the PCA model in the offline training step, SPE, $T^2$, and $\varphi$ index for the reconstruction error vector can be calculated using \ref{eq:SPE_ind}, \ref{eq:T2_ind}, and \ref{eq:combine_lim}. This study uses the $\varphi$ index for HIF detection since it can effectively combine the SPE and $T^2$ indices. If the combined index of $\mathbf{e}$ is above the control limit calculated in the offline training phase, it indicates that there are abnormal distortions that break the normal correlation structure in the cycle corresponding to the vector $\mathbf{x}$. To account for the noise and transient disturbances in the measurements, we use a counter to record the number of cycles with indices above the control limit. The counter is incremented when the combined index corresponding to one cycle exceeds the control limit and decreases if the index drops below the control limit. A trip signal is issued when the counter exceeds a predetermined threshold, which means the trip signal will be generated if the combined index consistently stays above the control limit.

\section{Evaluation}\label{sec:result}
\subsection{Dataset}
The dataset used in this study was collected during the testing and evaluation of ABB’s feeder protection system REF 550 \cite{ras:2007, ref550}. The measurements of three-phase voltages and currents were collected in a 4.16 kV distribution system near a hospital. High impedance faults in phase A were stages at about 12 miles from the hospital by dropping the conductor on four different surfaces: grass, water puddle, soil, and asphalt. In addition, the faults were created multiple times for each surface under different load conditions. In each case, a fault was introduced at around 100 seconds and lasted for 60 seconds before the conductor was lifted off the test surface. In addition to the fault cases, measurements of normal load were recorded. The number of samples per cycle $ts$ for all the measurements is 320. 

\begin{figure}[h]
  \centering
  \includegraphics[width=0.9\linewidth]{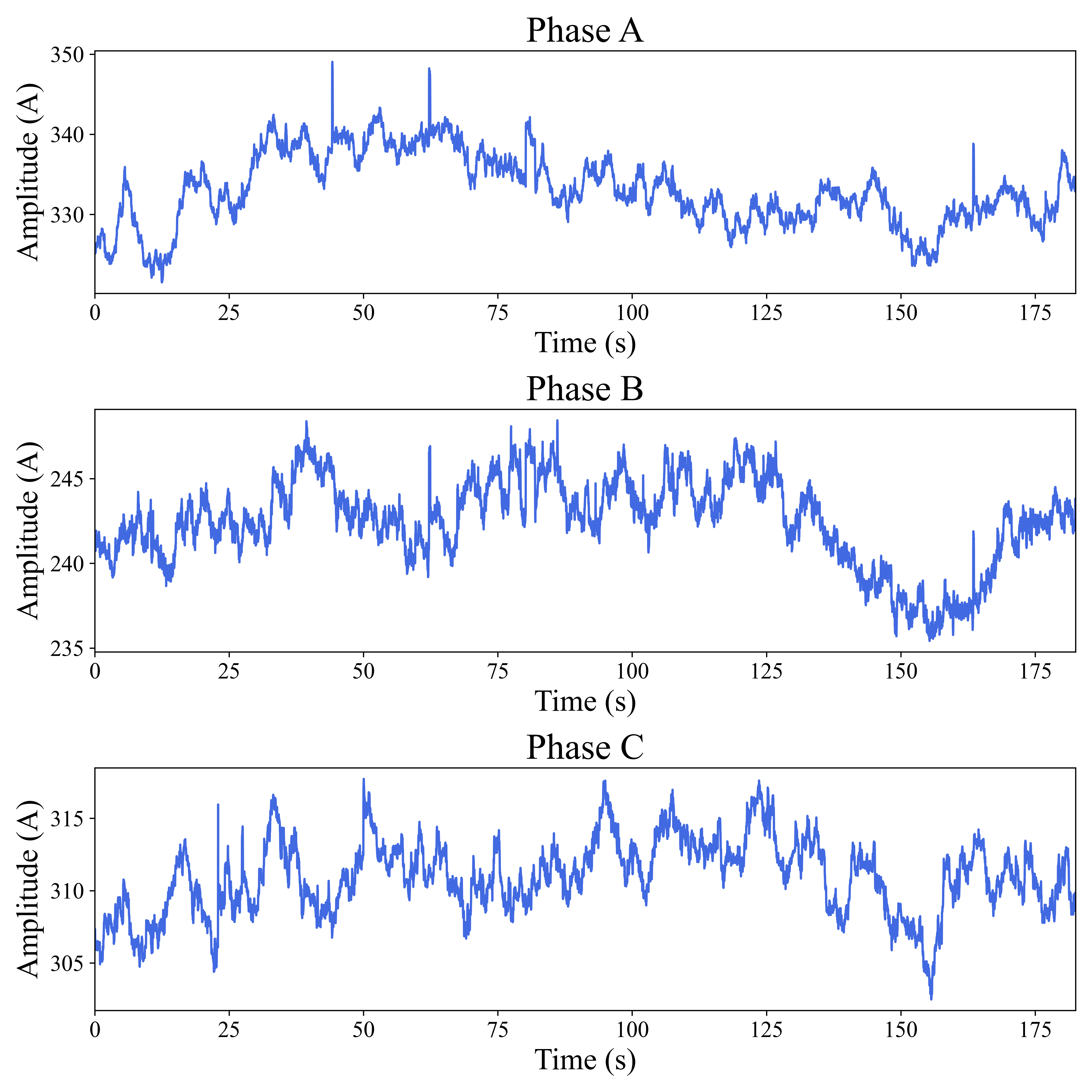}
\caption{RMS currents of normal load.}
\label{fig:rms}
\end{figure}

\begin{figure}[h]
  \centering
  \includegraphics[width=0.9\linewidth]{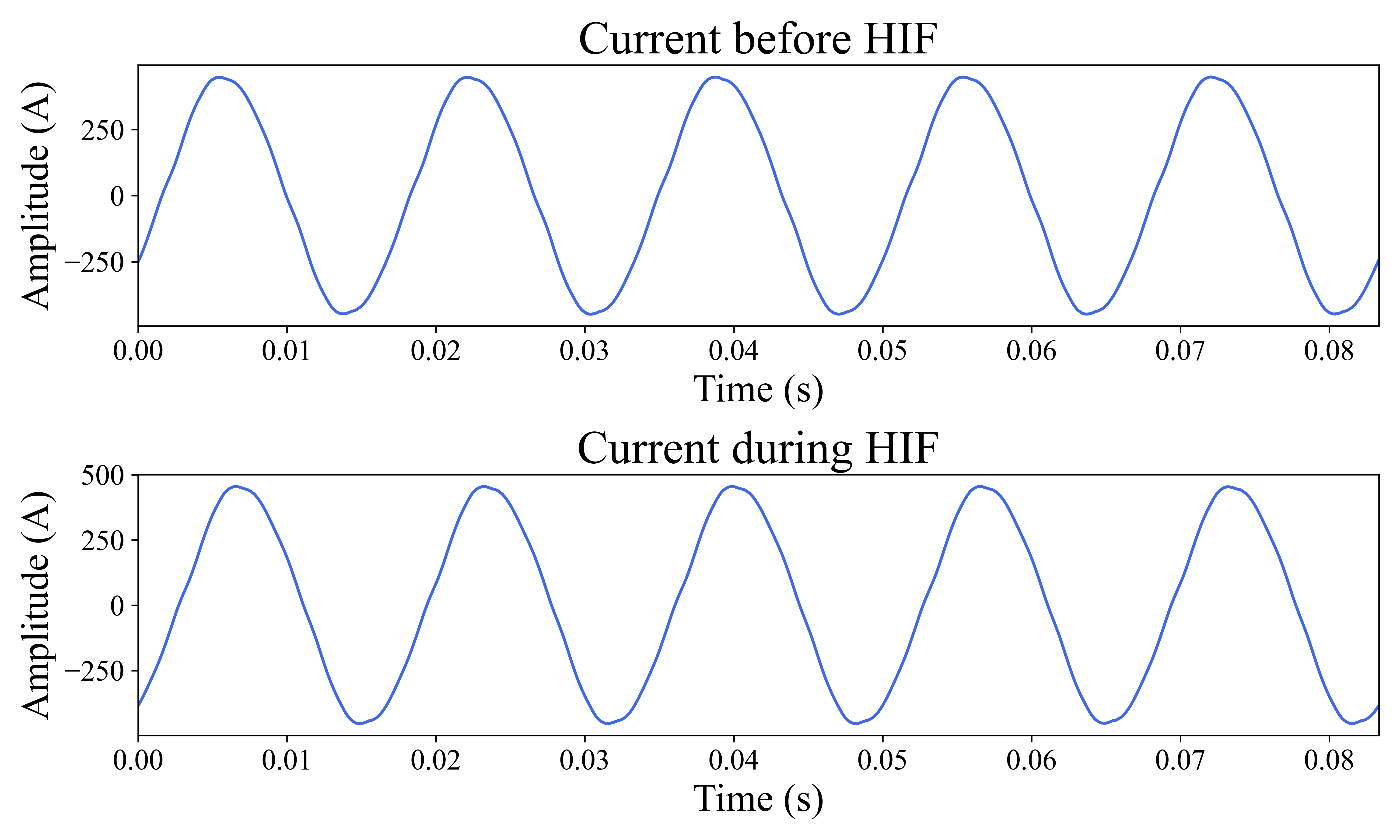}
\caption{Current waveform of the normal load and the waveform during the HIF.}
\label{fig:compare}
\end{figure}

Figure \ref{fig:rms} shows the root mean square (RMS) current waveforms of a section of normal loads. It can be observed that the variations in the load are dynamic and complex, with the occasional presents of spikes. In addition, the three phases are unbalanced with distinct patterns. Figure \ref{fig:compare} shows the comparison between the current waveform of the normal load and the waveform during HIF. Unlike the simulated cases used in previous publications \cite{rai:2021, wang_dehghanian:2020}, the load waveform is distorted and dynamic. As a result, distinguishing the HIF from the normal load is more challenging since the magnitudes of distortion in the two cases are similar.
\subsection{Results and Analysis}
The first step in implementing the proposed fault detection procedure is to augment the current waveforms to a data matrix. There are four load cases in the dataset. Three load cases containing around 580 seconds of measurements are used to train and validate the autoencoder model, and the last load case is left for testing. First, for each phase in each load case, the univariate current measurement is converted to a data matrix. Since the number of samples per cycle $ts$ is 320, the sampling gap $\Delta$ is selected to be 10, resulting in a data matrix consisting of 32 variables. As a result, each row in the matrix corresponds to the measurement sampled from one cycle. Next, all the data matrices formed from all three phases in three load cases are concatenated. After scaling each column of the concatenated data matrix to have values between 0 and 1, 80\% of the data is used for training, and 20\% is used for validation.

The autoencoder model used in this study has five layers. The dimension of the input and output layers are 32, and the dimension of the three hidden layers are 15, 10, and 15, respectively. The rectified linear unit (ReLU) is used as the activation function for the input and hidden layers. The model is trained using Adam optimizer in PyTorch with a learning rate of 0.001 is used to minimize the MSE loss. The autoencoder model is trained for 100 Epochs with a batch size of 32.

After the autoencoder model is trained using the normal load data, the reconstruction errors of the training and validation data are used to build a PCA model. The number of leading PCs $l$ is selected so that the first $l$ PCs captured 95\% of the variances,  and the confidence level $\alpha$ is chosen to be 99\%. 

\begin{figure}
  \centering
  \includegraphics[width=0.9\linewidth]{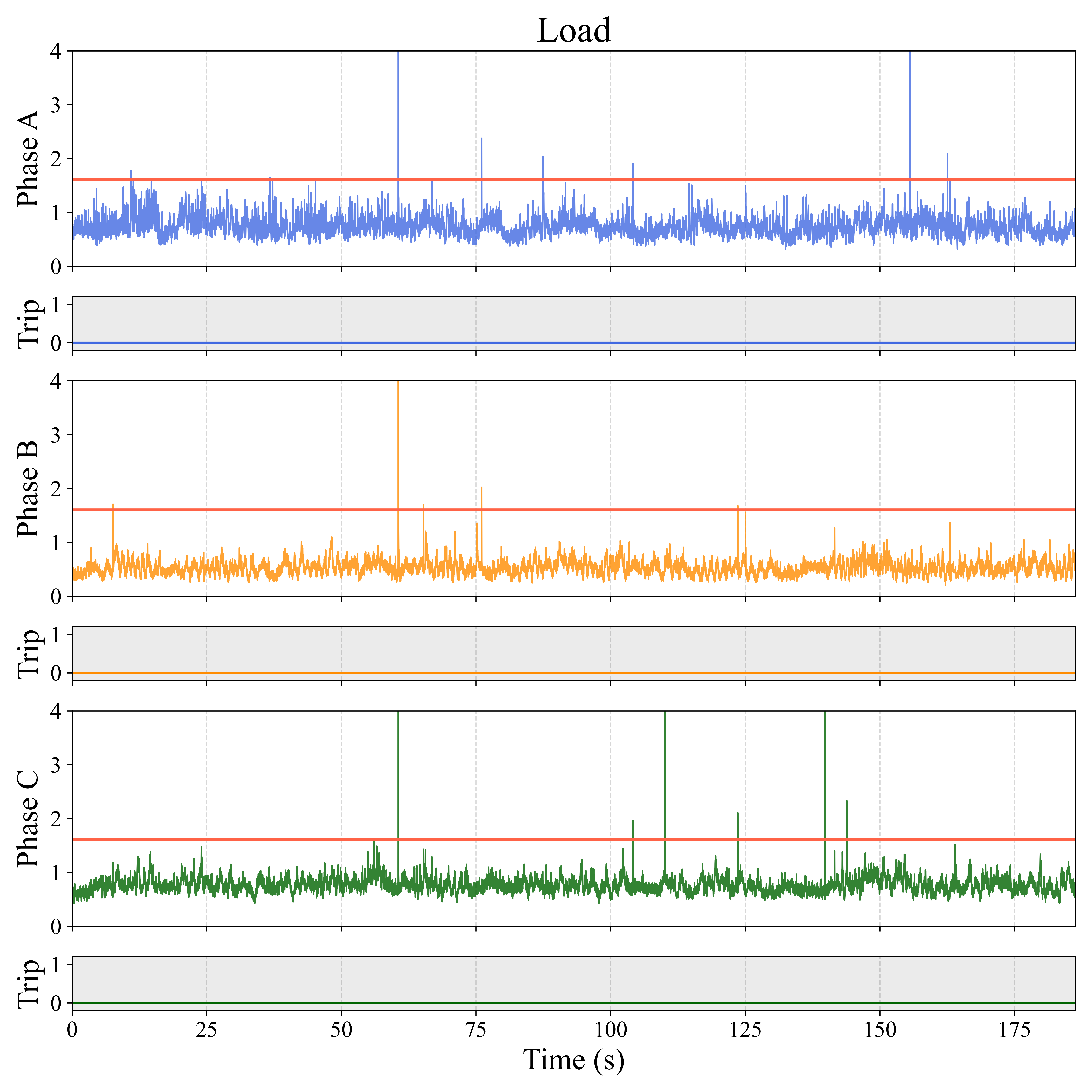}
\caption{Detection result of normal load.}
\label{fig:load}
\end{figure}

The autoencoder and PCA models are applied to the load and HIF cases staged on different surface types. The proposed HIF detection method is first applied to the load case that is not used during training to show that the proposed method does not generate false alarms for new load profiles. Figure \ref{fig:load} shows the combined indices and trip signals generated from a counter with a threshold of 60 for all three phases. It can be observed that most of the indices stay below the control limit, with a few outliers caused by spikes present in the current waveform. As a result, no trip signal is generated for all three phases, which is expected for the normal load.

\begin{figure}
  \centering
  \includegraphics[width=0.9\linewidth]{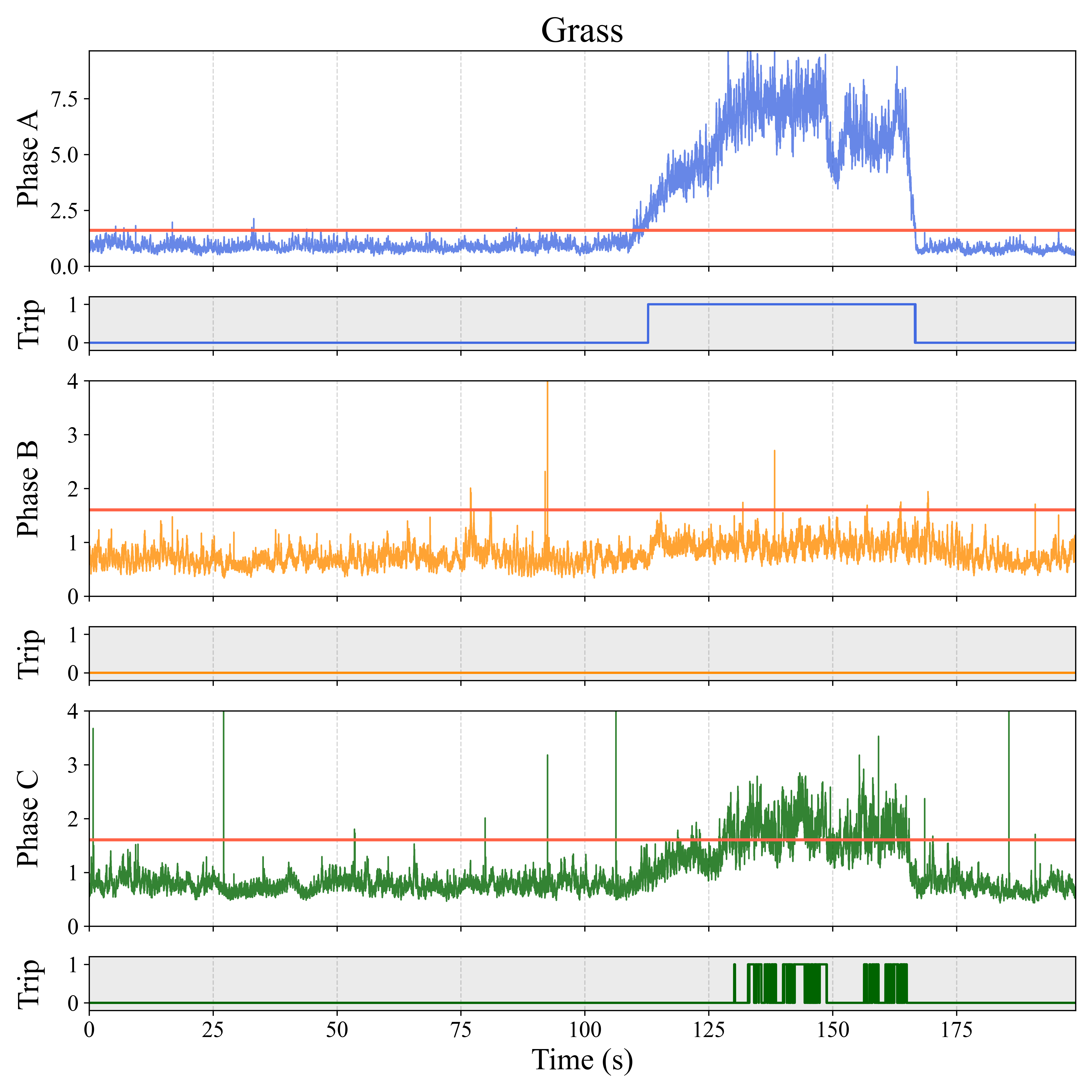}
\caption{Detection result of HIF on grass.}
\label{fig:grass}
\end{figure}

Three HIF cases were staged on the grass surface when ABB tested the REF 550 for HIF Detection, and the REF 550 failed to detect one of them. On the contrary, our proposed HIF detection can successfully detect all the HIF cases when the conductor of phase A contacts the grass. Figure \ref{fig:grass} shows the detection result of the proposed method for the case that REF 550 failed to detect. It can be observed that the combined index of phase A rises above the control limit after introducing HIF at around 100 seconds. The index stays above the control limit until the conductor is lifted off the grass at around 160 seconds. As a result, a trip signal is generated for phase A after the index stays above the control limit longer than 60 cycles. Phase C is also affected by the HIF. However, since the magnitude of its monitoring index is much smaller than phase A, HIF is determined to have occurred in phase A.  

\begin{figure}
  \centering
  \includegraphics[width=0.9\linewidth]{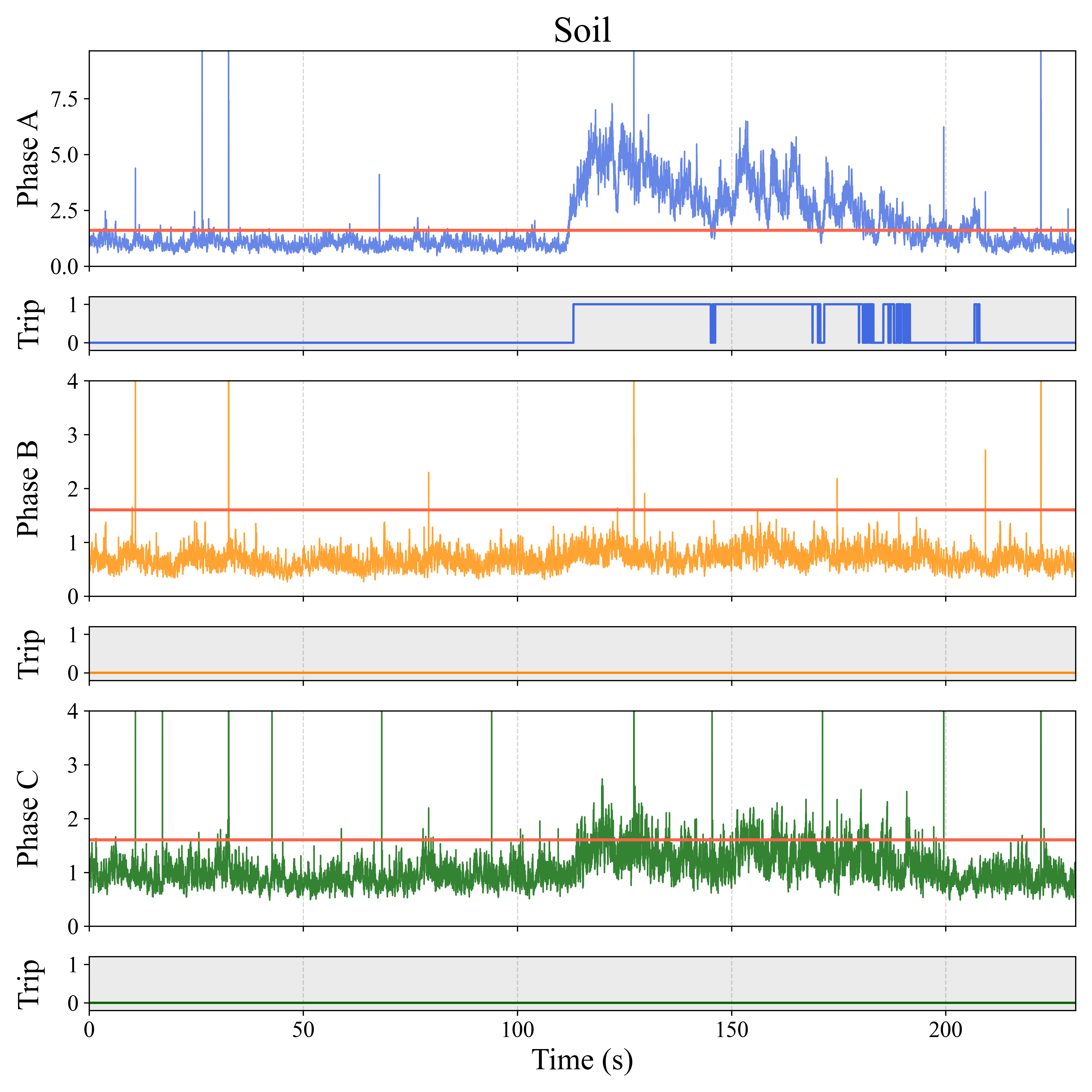}
\caption{Detection result of HIF on soil.}
\label{fig:Soil}
\end{figure}

In addition to the tests conducted on the grass surface, four HIF cases were staged by dropping the conductor of phase A on the soil surface. When these four cases were tested, the REF 550 detected three of them, and one was not detected. To compare our proposed HIF detection method, we apply the trained autoencoder and PCA models to these four cases, and the results show that all the HIFs can be detected. Figure \ref{fig:Soil} shows the detection result of the proposed method for the HIF case that REF 550 failed to detect. It can be seen that before the fault is introduced at around 100 seconds,  the indices for all three phases stay below the control limit, indicating that the current waveforms are normal and there is no fault. However, after the conductor of phase A contacts the soil, the monitoring index of phase A immediately rises and stays above the control limit. As a result, a trip signal is generated for phase A. Similar to the HIF cases staged on grass, phase C also shows minor abnormal distortions since the corresponding monitoring index oscillates around the control limit. However, the trip signal is not generated for phase C since the number of abnormal cycles does not reach the predefined threshold of 60. 

\begin{figure}
  \centering
  \includegraphics[width=0.9\linewidth]{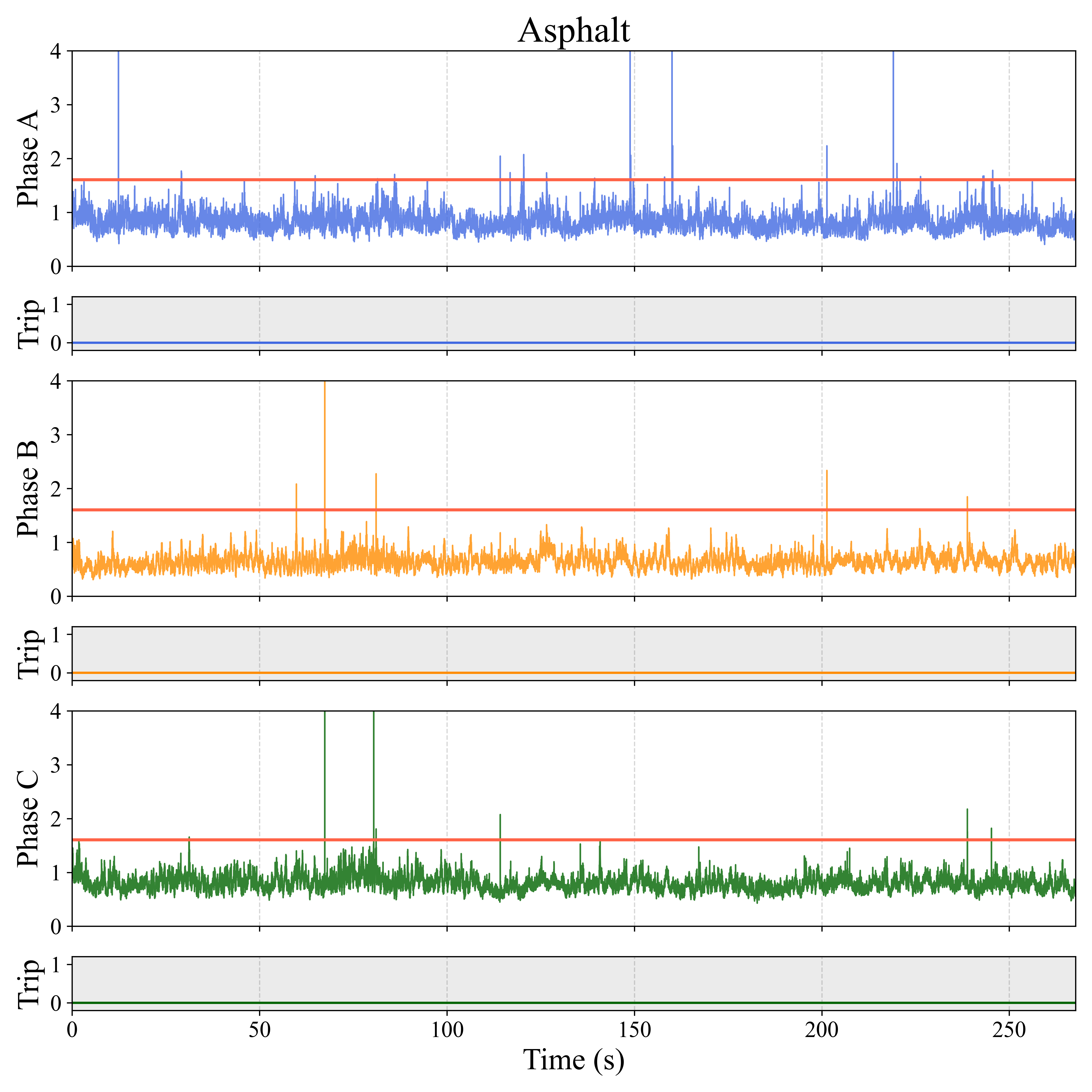}
\caption{Detection result of HIF on asphalt.}
\label{fig:asphalt}
\end{figure}

The dataset also contains measurements of high impedance faults on asphalt and puddle filled with drinkable water. During the testing, the REF 550 could not detect any HIF on asphalt and water puddle. Like the detection results from REF 550, our proposed method cannot detect any of these cases due to the near-infinite impedance conditions of the downed conductor test and the long distance between the fault location and where the measurements were taken. Figure \ref{fig:asphalt} shows the detection result of one of the HIF cases on asphalt in which no trip signal is generated since all the indices stay below the control limit. Even though the proposed method cannot detect the faults that occurred on near-infinite impedance surface types, no false alarms are generated during various load conditions before and after the HIFs in all the cases.

The comparison between the HIF detection results from the REF 550 and our proposed method can be summarized using the following metrics: accuracy (Acc), security (Sec), dependability (Dep), safety (Saf), and sensibility (Sen) \cite{ghaderi:2017}.
\begin{align}
    Acc & = \frac{TP + TN}{TP + TN + FP + FN} \times 100\% \\
    Sec & = \frac{TN}{TN + FP}\times100\% \\
    Dep & = \frac{TP}{TP + FN}\times100\% \\
    Saf & = \frac{TN}{TN + FN}\times100\% \\
    Sen & = \frac{TP}{TP + FP}\times100\% 
\end{align}
where true positives (TP) and true negatives (TN) are the numbers of the correctly detected
fault and normal load cases, and false negatives (FN) and false positives (FP) are the numbers of
the wrongly detected fault and load cases. We calculate the above metrics based on the detection results for all the cases in the entire dataset. The results are shown in Table \ref{tab:metric}.

Since the REF 550 and our proposed method can correctly identify the load conditions, they achieve 100\% dependability and security, indicating they are robust to faulty tripping. Furthermore, our proposed HIF detection method can correctly detect more HIF cases. As a result, our proposed HIF detection method shows improvement in the other metrics compared to the REF 550.
\begin{table}\label{tab:metric}
\caption{Comparison of REF 550 and proposed HIF detection method.}
\resizebox{0.5\textwidth}{!}{%
\begin{tabular}{|l|l|l|l|l|l|}
\hline
         & Acc    & Sec   & Dep    & Saf   & Sen    \\ \hline
REF 550  & 68.9\% & 100\% & 35.7\% & 62.5\% & 100\% \\ \hline
AE + PCA & 75.9\% & 100\% & 50\% & 68.2\% & 100\%   \\ \hline
\end{tabular}}
\end{table}
\section{Conclusion}\label{sec:conclusion}
This paper proposes an unsupervised HIF detection method based on the autoencoder and principal component analysis, which does not require measurements during HIFs. The proposed method first converts the univariate current measurement collected from one location in the grid into a data matrix. The data matrix is then used to train an autoencoder for extracting nonlinear features from the data matrix and capturing the correlations among variables in the data matrix. Finally, the PCA-based statistical monitoring technique is used to characterize the residuals of the normal load data from the AE model and establish thresholds based on various statistics. The proposed method detects high impedance faults by monitoring the deviation in the correlation structure of the augmented data matrix from the correlation learned by the AE from the normal loads. The proposed HIF detection method is applied to real data collected from a 4.16 kV distribution system which contains various normal load cases and HIF cases staged on four types of surfaces: grass, water puddle, soil, and asphalt. The detection results are compared with the results from the commercially available HIF detection solution REF 550, demonstrating that our proposed method outperforms REF 550 by detecting more HIF cases while not making false alarms during load conditions.

\bibliographystyle{IEEEtran}
\bibliography{references.bib}
\newpage

\end{document}